\documentclass[10pt,twocolumn,letterpaper]{article}

\usepackage[pagenumbers]{cvpr} 

\usepackage{graphicx}
\usepackage{amsmath}
\usepackage{amssymb}
\usepackage{booktabs}
\usepackage{times}
\usepackage{array,multirow}
\usepackage{color}
\usepackage{url}
\usepackage[table,xcdraw]{xcolor}
\usepackage[accsupp]{axessibility}

\usepackage[pagebackref,breaklinks,colorlinks]{hyperref}

\def\cvprPaperID{6294} 
\def\confName{CVPR}
\def\confYear{2023}

\begin{document}

\title{Text with Knowledge Graph Augmented Transformer for Video Captioning}

\author{Xin Gu$^{1,3}$, Guang Chen$^2$, Yufei Wang$^2$, Libo Zhang$^{3,1}$\thanks{Corresponding author:libo@iscas.ac.cn, Libo Zhang was supported by Youth Innovation Promotion Association, CAS (2020111). \protect\\ This work was done during internship at ByteDance Inc.}, Tiejian Luo$^1$, Longyin Wen$^2$\\
{$^1$University of Chinese Academy of Sciences, Beijing, China } \\
{$^2$ByteDance Inc., San Jose, USA} \\
{$^3$Institute of Software Chinese Academy of Sciences, Beijing, China}\\
{\tt \small guxin21@mails.ucas.edu.cn, guang.chen@bytedance.com, yufei.wang@bytedance.com}\\
{\tt \small libo@iscas.ac.cn, tjluo@ucas.ac.cn, longyin.wen@bytedance.com}
}

\maketitle

\begin{abstract}

Video captioning aims to describe the content of videos using natural language. Although significant progress has been made, there is still much room to improve the performance for real-world applications, mainly due to the long-tail words challenge. In this paper, we propose a text with knowledge graph augmented transformer (TextKG) for video captioning. Notably, TextKG is a two-stream transformer, formed by the external stream and internal stream. The external stream is designed to absorb additional knowledge, which models the interactions between the additional knowledge, \eg, pre-built knowledge graph, and the built-in information of videos, \eg, the salient object regions, speech transcripts, and video captions, to mitigate the long-tail words challenge. Meanwhile, the internal stream is designed to exploit the multi-modality information in videos (\eg, the appearance of video frames, speech transcripts, and video captions) to ensure the quality of caption results. In addition, the cross attention mechanism is also used in between the two streams for sharing information. In this way, the two streams can help each other for more accurate results. Extensive experiments conducted on four challenging video captioning datasets, \ie, YouCookII, ActivityNet Captions, MSR-VTT, and MSVD, demonstrate that the proposed method performs favorably against the state-of-the-art methods. Specifically, the proposed TextKG method outperforms the best published results by improving $18.7\%$ absolute CIDEr scores on the YouCookII dataset.

\end{abstract}

\section{Introduction}
Video captioning aims to generate a complete and natural sentence to describe video content, which attracts much attention in recent years. Generally, most existing methods \cite{HMN,SGN,Mart,DPC} require a large amount of paired video and description data for model training. Several datasets, such as YouCookII \cite{DBLP:conf/aaai/ZhouXC18}, and ActivityNet Captions \cite{DBLP:conf/iccv/KrishnaHRFN17} are constructed to promote the development of video captioning field. Meanwhile, some methods \cite{MVC,univl,Actbert,DECEM} also use the large-scale narrated video dataset HowTo100M \cite{DBLP:conf/iccv/MiechZATLS19} to pretrain the captioning model to further improve the accuracy. 

Although significant progress has been witnessed, it is still a challenge for video captioning methods to be applied in real applications, mainly due to the long-tail issues of words. Most existing methods \cite{DECEM,MVC,Actbert,univl} attempt to design powerful neural networks, trained on the large-scale video-text datasets to make the network learn the relations between video appearances and descriptions. However, it is pretty tough for the networks to accurately predict the objects, properties, or behaviors that are infrequently or never appearing in training data. Some methods \cite{image,bad} attempt to use knowledge graph to exploit the relations between objects for long-tail challenge in image or video captioning, which produces promising results. 

\begin{figure*}[t]
 \centering
 \includegraphics[trim=53 126 175 40, clip, width=0.95\linewidth]{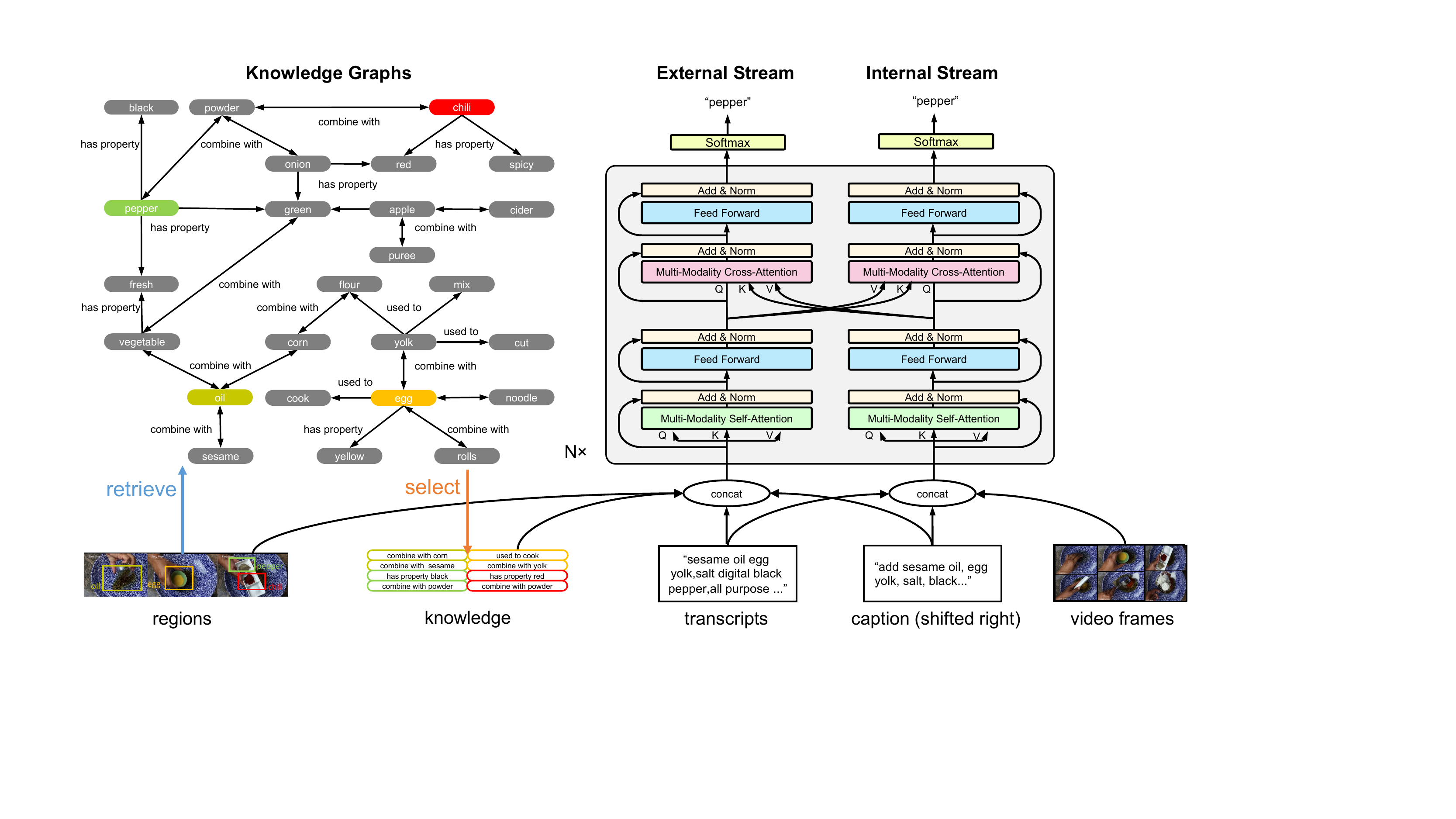}
 \caption{The architecture of our TextKG method, which is formed the external and internal streams. Each stream is stacked with $N$ sets of multi-modality self-attention and cross-attention modules.The cross attention modules are designed to align tokens in different modalities.}
 \label{fig:framework}
\vspace{-3mm} 
\end{figure*}

In this paper, we present a text with knowledge graph augmented transformer (TextKG), which integrates additional knowledge in knowledge graph and exploits the multi-modality information in videos to mitigate the long-tail words challenge. TextKG is a two-stream transformer, formed by the external stream and internal stream. The external stream is used to absorb additional knowledge to help mitigate long-tail words challenge by modeling the interactions between the additional knowledge in pre-built knowledge graph, and the built-in information of videos, such as the salient object regions in each frame, speech transcripts, and video captions. Specifically, the information is first retrieved from the pre-built knowledge graphs based on the detected salient objects. 
After that, we combine the features of the retrieved information, the appearance features of detected salient objects,  the features of speech transcripts and captions, then feed them into the external stream of TextKG to model the interactions. The internal stream is designed to exploit the multi-modality information in videos, such as the appearance of video frames, speech transcripts and video captions, which can ensure the quality of caption results. To share information between two streams, the cross attention mechanism is introduced. In this way, the two streams can obtain the required modal information from each other for generating more accurate results. The architecture of the proposed method is shown in Figure \ref{fig:framework}.

Several experiments conducted on four challenging datasets, \ie, YouCookII \cite{DBLP:conf/aaai/ZhouXC18}, ActivityNet Captions \cite{DBLP:conf/iccv/KrishnaHRFN17}, MSR-VTT \cite{DBLP:conf/cvpr/XuMYR16}, and MSVD \cite{DBLP:conf/acl/ChenD11} demonstrate that the proposed method performs favorably against the state-of-the-art methods. Notably, our TextKG method outperforms the best published results by improving $18.7\%$ and  $3.2\%$ absolute CIDEr scores in the paragraph-level evalution mode on the YouCookII and Activity-Net Captions datasets.

\section{Related Work}
{\noindent {\bf Video captioning}} attracts much attention of researchers in recent years. The best practice has been achieved by attention-based methods, which attempts to associate visual components with sentences in videos. Some of them focus on designing powerful network architectures. VLM \cite{VLM} and VideoBERT \cite{VideoBERT} take the visual and text modalities as input, and use a shared transformer to construct a task-agnostic video-language model. ViLBERT \cite{Vilbert} processes visual and linguistic information separately with two parallel streams, and then use the attention mechanism to model the interactions between visual and language features. Instead of using the separate encoder-decoder architecture, MART \cite{Mart} designs a shared encoder-decoder network and augments it with the memory module. ActBert \cite{Actbert} uses local regional features to learn better visual-language alignment. WLT \cite{wlt} takes audio features as an additional input, and uses context fusion to generate multimodal features. 

Meanwhile, some methods \cite{image,bad,tkg, BEIC, ICiek, DST} focus on exploiting prior knowledge to provide semantic correlations and constraints between objects for image or video captioning, producing promising results. ORG-TRL \cite{ORG-TRL} uses the knowledge information in the language model (BERT) to provide
candidate words for video captioning. In contrast, we propose a two-stream transformer for video captioning, with the internal stream used to exploit multi-modality information in videos, and the external stream used to model the interactions between the additional knowledge and the built-in information of videos. These two streams use the cross-attention mechanism to share information in different modalities for generating more accurate results.

{\noindent {\bf Vision-and-language representation learning}} is a hot topic in recent years. ViLBERT \cite{Vilbert}, LXMERT \cite{LXMERT}, UNITER \cite{UNITER}, UNIMO \cite{UNIMO} and Unified-VL \cite{Unified-VL} learn the representations between image and text, while Univl \cite{univl}, VideoBERT \cite{VideoBERT}, ActBERT \cite{Actbert} and MV-GPT \cite{MVC} learn the representations between videos and transcripts. Notably, most of these methods attempt to learn powerful vision-and-language representations by pre-training the models on the large-scale datasets, \eg, Howto100M \cite{howto100} and WebVid-2M \cite{Frozen}, and then finetune them on downstream tasks such as video captioning, video-text retrieval and visual question answering. In contrast, our TextKG method uses the speech transcripts as the text to model the visual and linguistic representations and integrate the additional knowledge in knowledge graph to mitigate long-tail words challenge in video captioning.

{\noindent {\bf Knowledge graph in NLP.}}
Knowledge graph is an useful tool to indicate the real-world entities and their relations, which provides rich structured knowledge facts for language modeling. Large-scale knowledge graphs are used to train knowledge enhanced language models for various natural language processing (NLP) tasks. CoLAKE \cite{CoLake} proposes to inject the knowledge context of an entity, and to jointly learn the contextualized representation for both language and knowledge by a unified structure. ERNIE \cite{ERNIE} enhances BERT architecture to better integrate the knowledge information and textual information. KEPLER \cite{KEPLER} not only improves language models by integrating factual knowledge, but also generates text-enhanced knowledge representation. JAKET \cite{JAKET} proposes a joint pre-training framework to model knowledge graph and language simultaneously. Inspired by CoLAKE, our method jointly learns the representations of vision, language and knowledge, and enhances the joint visual-language representations by retrieving relevant knowledge in knowledge graphs.

\section{Our Approach}
As mentioned above, our method aims to integrate additional knowledge using knowledge graph and exploits the multi-modality information in videos to mitigate the long-tail words challenge in the field of video captioning. We design a text with knowledge graph augmented transformer, \ie, a two-stream transformer formed by the external and internal streams. Both streams are constructed by $N$ sets of alternately stacked self-attention and cross-attention blocks. Specifically, we first use a detector to generate the salient object regions in each video frame, and use the automatic speech recognition (ASR) method \cite{asr} to generate the speech transcripts in videos. After that, the class labels of detected salient objects are used to retrieve the prior knowledge in knowledge graphs. Then, the appearance embeddings of detected salient objects and video frames, and embeddings of retrieved prior knowledge, speech transcripts, and predicted captions are fed into the two-stream transformer to generate subsequent words of the predicted captions. The overall architecture of the proposed TextKG is shown in Figure \ref{fig:framework}.

\subsection{Two-Stream Transformer}
As described above, our two-stream transformer is formed by the external stream and internal stream. Several self-attention and cross-attention blocks are interleaved to construct the two streams.

Let $\{ {\it y}_{1}, {\it y}_{2}, \cdots, {\it y}_{l} \}$ be the predicted captions, where ${\it y}_{i}$ is the index of the $i$-th word in the dictionary, $\{{\it p}_{1}, {\it p}_{2}, \cdots, {\it p}_{l}\}$ to be the predicted probabilities of words in the dictionary, where ${\it p}_{i}$ is the probability of the $i$-th word. The index of the $i$-th word ${\it y}_{i}$ is computed as ${\it y}_{i}=\mathop{\arg\max}_{i}{\it p}_{i}$. Meanwhile, let $z^{\text{ext}}_{i}$ and $z^{\text{int}}_{i}$ to be the output probabilities of the external and internal streams of the $i$-th word. Thus, we have ${\it p}_{i} = \omega_1 z^{\text{ext}}_{i} + \omega_2 z^{\text{int}}_i$, where $\omega_1$ and $\omega_2$ are the hyperparameters used to balance the outputs of the two-streams. In this way, ${\it y}_{i}$ is computed as 
\begin{equation}
    \begin{array}{ll}
{\it y}_{i}=\mathop{\arg\max}_{i}(\omega_1 z^{\text{ext}}_{i} + \omega_2 z^{\text{int}}_i). 
    \end{array}
\end{equation}
In the following sections, we will describe each module in our two-stream transformer in more details.

{\noindent {\bf Multi-modality Self-Attention Module}.}
As shown in Figure \ref{fig:framework}, we use the self-attention module \cite{DBLP:conf/nips/VaswaniSPUJGKP17} in both the external and internal streams to model the interactions among multi-modality information. Specifically, for the external stream, the concatenation of feature embeddings of detected objects ${\cal F}_{\text{r}}$, retrieved prior knowledge ${\cal F}_{\text{k}}$, speech transcripts ${\cal F}_{\text{s}}$, and predicted video captions ${\cal F}_{\text{c}}$ are fed into the self-attention module to model the interactions, \ie, ${\cal X}_{\text{ext}}=\textnormal{Concat}({\cal F}_{\text{r}}, {\cal F}_{\text{k}}, {\cal F}_{\text{s}}, {\cal F}_{\text{c}})$. Meanwhile, for the internal stream, the concatenation of feature embeddings of speech transcripts ${\cal F}_{\text{s}}$, predicted video captions ${\cal F}_{\text{c}}$, and video frame ${\cal F}_{\text{v}}$ are fed into the self-attention model to exploit the interactions, \ie, ${\cal X}_{\text{int}}=\textnormal{Concat}({\cal F}_{\text{s}}, {\cal F}_{\text{c}}, {\cal F}_{\text{v}})$. The self-attention module computes the interactions as follows,
\begin{equation}
\begin{array}{ll}
\Phi(Q,K,V)={\textnormal{softmax}}({QK^T \over \sqrt{d}} + {\it M})V,
\end{array}
\label{equ:self-attention}
\end{equation}
where $d$ is the feature dimension of queries $Q$ and keys $K$. We have $Q={\cal X}_{(\cdot)}{\it W}^{\textnormal{Q}}$, $K={\cal X}_{(\cdot)}{\it W}^{\textnormal{K}}$, $V={\cal X}_{(\cdot)}{\it W}^{\textnormal{V}}$, where ${\it W}^{\textnormal{Q}}$, ${\it W}^{\textnormal{K}}$, and ${\it W}^{\textnormal{V}}$ are learnable parameters, and ${\cal X}_{(\cdot)}\in\{{\cal X}_{\text{ext}}, {\cal X}_{\text{int}}\}$ are the concatenated feature embeddings for the external and internal streams, respectively. 

Following \cite{Mart}, we introduce a mask matrix ${\it M}$ in attention function \eqref{equ:self-attention} of both the external and internal streams to prevent the model from seeing future words. That is, for the $i$-th caption token, we set ${\it M}_{i, j} = 0$ for $j=1,\cdots, i$, and set ${\it M}_{i, j}=-\inf$ for $j>i$. Meanwhile, for the external stream, we set ${\it M}_{i, j}=-\inf$ to prevent the associations between irrelevant retrieved prior knowledge and the detected salient object regions.

{\noindent {\bf Multi-modality Cross-Attention Module.}}
Besides self-attention module, we use the cross-attention module to align the interactions modeled by both the external and internal streams. Specifically, we compute the affinity matrix to guide the alignments between the modeled interactions by injecting the information. Similar to the self-attention module, we also introduce the mask matrix into the attention function to compute the cross-attention. Intuitively, the retrieved prior knowledge should not have direct effect on the predicted captions, but through the detected salient objects. Thus, we set the corresponding elements to $-\inf$ to prevent the retrieved prior knowledge directly influencing the predicted captions. 

\subsection{Optimization}
The cross entropy loss function is used to guide the training of our TextKG method. Given the ground-truth indices of previous $(i-1)$ words and the concatenated of two-streams ${\cal X}_{\text{ext}}$ and ${\cal X}_{\text{int}}$, we can get the predictions of the current $i$-th word. After that, the training loss of our method is computed as 
\begin{equation}
    \begin{array}{ll}
    {\cal L}=-\sum_{i=1}^{l}(\lambda_1\log{\it z}^{\text{ext}}_{i}+
 \lambda_2\log{\it z}^{\text{int}}_{i}),
    \end{array}
\end{equation}
where ${\it z}^{\text{ext}}_{i}$ and ${\it z}^{\text{int}}_{i}$ are the $i$-th output word of the external and internal streams, $\lambda_1$ and $\lambda_2$ are the preset parameters used to balance the two streams, and $l$ is the total length of predicted captions. The Adam optimization algorithm \cite{DBLP:journals/corr/KingmaB14} with an initial learning rate of $1e-4$, $\beta_1=0.9$, $\beta_2=0.999$ and $L2$ weight decay of $0.01$ is used to train the model. The Warmup strategy is adopted in training phase by linearly increasing the learning rate from $0$ to the initial learning rate for the first $10\%$ of training epochs, and linearly decreasing to $0$ for the remaining $90\%$ of epochs.

\subsection{Multimodal Tokens}
The appearance embeddings of detected salient objects and video frames, and embeddings of retrieved prior, speech transcripts, and predicted captions are fed into the two-stream transformer to generate subsequent captions after tokenization. We will describe the feature extraction and tokenization processes of the aforementioned embeddings.

{\noindent \textbf{Video tokens}} encode the appearance and motion information of video frames. Specifically, the aligned appearance and motion features form the video tokens. In each second of the video, we sample $2$ frames to extract features. Following \cite{Mart}, we use the ``Flatten-673'' layer in ResNet-200 \cite{resnet} to extract appearance features, and the ``global pool'' layer in BN-Inception \cite{BN} to extract the motion features on the YouCookII \cite{DBLP:conf/aaai/ZhouXC18} and ActivityNet Captions \cite{DBLP:conf/iccv/KrishnaHRFN17} datasets. Meanwhile, for fair comparison, we use the InceptionResNetV2 \cite{incev2} and C3D \cite{c3d} models to extract appearance and motion features on the MSR-VTT \cite{DBLP:conf/cvpr/XuMYR16} and MSVD \cite{DBLP:conf/acl/ChenD11} datasets, following the methods \cite{HMN,ORG-TRL,SAAT}.

{\noindent \textbf{Region tokens}} are used to describe the visual information of the detected salient objects. We use the Faster R-CNN method \cite{fasterrcnn} to extract visual features for detected object regions, which is pretrained on the Visual Genome dataset \cite{VG}, and the ${\it N}_{\text{r}}$ detected objects with highest confidence score in each frame are reserved for video captioning.

{\noindent \textbf{Transcript tokens}} are used to encode the content of speech transcripts, which contains some key information in videos. The GloVe model \cite{Glove} is used to extract features of each word in transcripts to form the transcript tokens.

{\noindent \textbf{Caption tokens}} encode the information of predicted captions. Similar to transcript tokens, GloVe \cite{Glove} is adopted to extract features of each word in the generated captions.

{\noindent \textbf{Knowledge tokens}} model the content of retrieved knowledge, which is important to mitigate the long-tail words challenge in video captioning. Our TextKG method uses a triple structure to encode the retrieved knowledge, formed by two object items and the relations between them, \ie, $(\alpha_{\text{head}}, \alpha_{\text{tail}}, \rho_{\text{rel}})$, where $\alpha_{\text{head}}$ and $\alpha_{\text{tail}}$ are the two nodes in knowledge graph, and $\rho_{\text{rel}}$ indicates the edge encoding the relations between them. For example, $\alpha_{\text{head}}=$``\texttt{knife}'', $\alpha_{\text{tail}}=$``\texttt{hard}'', and $\rho_{\text{rel}}=$``\texttt{has property}''. For each detected object $\alpha_{\text{head}}$, we aim to integrate prior context information for more accurate caption results. That is, we use the GloVe model to extract linguistic features of the node $\alpha_{\text{tail}}$ and set the learnable embeddings for the edge $\rho_{\text{rel}}$. After that, we compute the summation of them to get the knowledge embeddings of the knowledge tokens.  

For each detected object, we use its category name to retrieve knowledge (\ie, triples containing the category name of the detected object) from the knowledge graph. Notably, only a portion of the retrieved knowledge are highly related to the video. The rests are potentially harmful to the performance of caption generation as it may distract the training and inference and also increase computational cost. For example, to generate caption of a cooking tutorial video, the knowledge ``\texttt{knife - used to -cut}'' is more useful rather than ``\texttt{knife - has property - hard}''. Because cutting food may be an important procedure for food preparation. Thus, we design a primary rank mechanism to sort the retrieved knowledge based on their semantic similarity to the video. The cosine similarity between the emebeddings of knowledge tokens and video transcripts is computed as the semantic similarity for ranking. The pretrained SBERT \cite{sbert} model is used to extract feature embeddings of transcripts. Finally, ${\it N}_{\text{k}}$ knowledge items with the highest similarity scores are reserved for video captioning.

\subsection{Knowledge Graphs Construction}
As mentioned above, we use the knowledge graph to retain the prior knowledge. The nodes in the graph could corresponding to a noun, an adjective, a verb or an adverb, etc. The edges in the graph are used to encode the relations between different nodes. Apparently, it is important to include all key information of videos in knowledge graphs for accurate caption results. Thus, besides general knowledge graph covering most key knowledge in general scenarios, we also introduce the specific knowledge graph for each dataset covering the key knowledge in specific scenarios.

{\noindent \textbf{General knowledge graph}} is designed to include most key information in general scenarios that we are interested, such as cooking  and activity. Specifically, the general knowledge graph is exported from the public available giant knowledge graph ConceptNet \cite{ConceptNet} by extracting key words\footnote{The key words are exported from the speech transcripts and annotated ground-truth captions in the training sets of the caption datasets.} in ConceptNet with the connected edges and neighboring nodes.

{\noindent \textbf{Specific knowledge graph.}} Besides the general knowledge graph, we also construct the specific knowledge graph to cover key information in specific scenarios. We believe that the speech transcripts of videos contain most of the crucial information and use the speech transcripts as the source to construct the specific knowledge graph. We first use the automatic speech recognition (ASR) model \cite{asr} to convert the speech in all videos to transcripts. After that, we use the Stanford NLP library\footnote{https://stanfordnlp.github.io/CoreNLP/} to analyze the components in each sentence in transcripts to form a structured grammer tree. We collect the ``adjective and noun'', ``noun and noun'' and ``adverb and verb'' phrases from the grammar trees to construct the specific knowledge graph.

\section{Experiments}
\subsection{Datasets}
{\noindent \textbf{YoucookII}} contains $2,000$ long untrimmed videos from $89$ cooking recipes, including $1,333$ videos for training, $457$ videos for testing. The average length of videos is $5.3$ minutes and each video is divided into $7.7$ video clips on average with the manually labeled caption sentences.

{\noindent \textbf{ActivityNet Captions}} is a large-scale challenging dataset for video captioning, with $10,024$ videos for training, $4,926$ for validation, and $5,044$ for testing. The average length of each video is $2.1$ minutes and is divided into $3.65$ clips on average. Each clip is assigned with a manually labeled caption sentence. Since the testing set is not publicly available, following MART \cite{Mart}, we split the validation set into two subsets, \ie, ae-val with $2,460$ videos for validation and ae-test with $2,457$ videos for testing. 

{\noindent \textbf{MSR-VTT}} includes $10,000$ video clips with $15$ seconds on average. Each video clip contains $20$ human annotated captions and the average length of each sentence is $9.28$ words. Following \cite{HMN,MGRMP,SGN}, $6,513$, $497$, and $2,990$ videos are used for training, validation, and testing, respectively. 

{\noindent \textbf{MSVD}} consists of $1970$ video clips collected from YouTube and each clip is approximately $10$ seconds long. Each clip is labeled with $40$ sentences. Following \cite{HMN,MGRMP,SGN}, we split the dataset into $1,200$, $100$, and $670$ video clips for training, validation and testing, respectively.

\subsection{Evaluation metrics}
To compare the performance of the proposed method with other methods, five model-free automatic evaluation metrics are used, including BLEU@4 \cite{BLEU} (abbreviated to B) that are precision-based metric, METEOR \cite{METEOR} (abbreviated to M) that computes sentence-level scores, CIDEr \cite{CIDEr} (abbreviated to C) that is consensus-based metric, ROUGE \cite{ROUGE} (abbreviated to R) that uses longest common subsequence to compute the similarities between sentences, and Rep@4 \cite{Re} (abbreviated to Rep) that computes the redundancy score for every captions. Lower Rep indicates better model. The CIDEr score is used as the primary metric in evaluation. Notably, both micro-level and paragraph-level evaluation modes are used. The micro-level mode evaluates the predicted captions of each video clip independently, while the paragraph-level mode considers the relations between captions of different clips. Following \cite{Mart,COOT}, the paragraph-level evaluation mode is used in the YouCookII and ActivityNet datasets, which contains multiple clips in each video.

\subsection{Implementation Details}
We sample every $2$ frames per second in each video and use the pre-trained models to extract video features. For the YouCookII and ActivityNet Captions datasets, similar to \cite{Mart}, the ResNet-200 and BN-Inception models are used to extract the appearance and motion features, respectively. Following \cite{HMN,ORG-TRL,SAAT}, the InceptionResNetV2 and C3D models are used to extract the appearance and motion features for the MSR-VTT and MSVD dataset, respectively. Meanwhile, the Faster R-CNN method using ResNet-101 as the backbone is adopted to detect object in each video frame, which is trained on the Visual Genome dataset \cite{VG}. Notably, only $N_{\text{r}}=6$ classes of the detected objects with top scores are used to form region features in each frame. 

The caption sentences, speech transcripts, and retrieved knowledge in knowledge graph are split into individual words using the NLTK toolkit\footnote{https://github.com/nltk/nltk}, and the GloVe \cite{Glove} algorithm is used to extract the word embeddings for each word to construct the transcript, caption, and knowledge tokens. The dimension of word embeddings is set to $300$. The captions are truncated to $20$ words and the maximum text length of transcripts is set to $300$ words. For each detected object, we retrieve up to $N_{k}=5$ pieces of knowledge items in knowledge graph depending on the relevant scores.

Notably, the proposed method is implemented using Pytorch. $3$ blocks of multi-modality self-attention and cross-attention modules are used in the MSR-VTT \cite{DBLP:conf/cvpr/XuMYR16} and MSVD \cite{DBLP:conf/acl/ChenD11} datasets, and $2$ are used in the YouCookII \cite{DBLP:conf/aaai/ZhouXC18} and ActivityNet Captions \cite{DBLP:conf/iccv/KrishnaHRFN17} datasets, to form the two-stream transformer, respectively. The feature dimension in each block is set to $768$, and the number of heads in multi-head architecture is set to $12$. The proposed method is trained using the Adam algorithm \cite{DBLP:journals/corr/KingmaB14} with the initial learning rate of $1e-4$. The batch size in training phase is set to $6$. The hyper-parameters $\omega_{1}$, $\omega_{2}$, $\lambda_1$, and $\lambda_2$ are set to $0.8$, $0.2$, $0.5$ and $0.5$, empirically.

\subsection{Comparison to the State of the Art Methods}
{\noindent \textbf{YouCookII dataset}.} To validate the effectiveness of the proposed method, we compare it to the state-of-the-art (SOTA) methods using the paragraph-level evaluation mode, reported in Table \ref{tab-youcookii-paragraph}. As shown in Table \ref{tab-youcookii-paragraph}, our TextKG method achieves the best results on the YouCookII dataset. Specifically, TextKG improves $18.7\%$ absolute CIDEr score compared to the SOTA method COOT \cite{COOT}. COOT \cite{COOT} focuses on leveraging the hierarchy information and modeling the interactions between different levels granularity and different modalities, such as frames and words, clip and sentences, and videos and paragraphs, which is pre-trained on the large-scale HowTo100M \cite{howto100} dataset. In contrast, our TextKG method aims to exploit additional knowledge in knowledge graph and multi-modality information in videos to improve the caption results.

Beside the paragraph-level evaluation mode, we also report the evaluation results based on the micro-level mode on the YouCookII dataset in Table \ref{tab-youcookii-micro}. As shown in Table \ref{tab-youcookii-micro}, TextKG significantly outperforms other methods, including Univl \cite{univl} and AT \cite{AT}, both of which use transcript information to enhance model performance by directly concatenating visual and transcript embedding. We argue that with the use of knowledge graphs, TextKG gives a better understanding of the content included in the speech transcripts and thus yields favorable results.

\begin{table}[t]
\centering
\setlength{\tabcolsep}{9.5pt}
\small{
\begin{tabular}{c|cccc}
\toprule
Method &B &M &C &Rep \\
\midrule
Van-Trans \cite{MaskedTrans} & 7.6 & 15.7 & 32.3 & 7.8 \\
Trans-XL \cite{Transformer-XL} & 6.6 & 14.8 & 26.4 & 6.3 \\
Trans-XLRG \cite{Transformer-XLRG} & 6.6 & 14.7 & 25.9 & 6.0 \\
MART \cite{Mart} & 8.0 & 16.0 & 35.7 & 4.4 \\
Van-Trans+COOT \cite{COOT} & 11.1 & 19.8 & 55.6 & 5.7\\
COOT \cite{COOT} & 11.3 & 19.9 & 57.2 & 6.7 \\ \hline
TextKG & \textbf{14.0} & \textbf{22.1} & \textbf{75.9} & \textbf{2.8} \\ \hline
Human & - & - & - & 1.0 \\
\bottomrule
\end{tabular}}
\caption{Evaluation results on the YouCookII val subset in the paragraph-level evaluation mode.}
\label{tab-youcookii-paragraph}
\vspace{-4mm}
\end{table}

\begin{table}[t]
\centering
\small{
\begin{tabular}{c|c|ccccc}
\toprule
Method & Input &B &M &R &C \\ \hline
Masked Trans \cite{MaskedTrans} & V & 3.8 & 11.6 & 27.4 & 38 \\
S3D \cite{S3D} & V & 3.2 & 9.5 & 26.1 & 31 \\
VideoAsMT \cite{VideoASMT} & V & 5.3 & 13.4 & - & - \\
SwinBERT \cite{Swinbert} & V & 9 & 15.6 & 37.3 & 109 \\
VideoBERT \cite{VideoBERT}  & V & 4.0 & 11.0 & 27.5 & 49 \\
VideoBERT+S3D \cite{VideoBERT} & V & 4.3 & 11.9 & 28.8 & 50 \\
ActBERT \cite{Actbert} & V & 5.4 & 14.3 & 30.6 & 65 \\
AT \cite{AT}& V+S & 9.0 & 17.8 & 36.7 & 112 \\
DPC \cite{DPC}& V+S & 2.8 & 18.1 & - & - \\
VALUE \cite{value} &  V+S & 12.4 & \textbf{18.8} & \textbf{40.4} & 130 \\
Univl \cite{univl} & V+S & 9.5 & 16.3 & 37.4 & 115 \\
MV-GPT \cite{MVC}& V+S & \textbf{13.3} & 17.6  & 35.5 & 103 \\ \hline
TextKG & V+S & 11.7 &18.4 &40.2 &\textbf{133} \\
\bottomrule
\end{tabular}}
\caption{Evaluation results on the YouCookII val subset in the micro-level evaluation mode. ‘V’ indicates the methods use video appearance information, and ‘S’ indicates the methods use the speech information.}
\label{tab-youcookii-micro}
\end{table}

{\noindent \textbf{ActivityNet Caption dataset.}} We also evaluate our TextKG method on the challenging ActivityNet Caption dataset in Table \ref{tab-activity-paragraph}. As shown in Table \ref{tab-activity-paragraph}, TextKG achieves the best results on $3$ out of $4$ metrics, \ie, BLEU, METOR, and CIDEr. The ActivityNet Caption dataset \cite{DBLP:conf/iccv/KrishnaHRFN17} contains a series of complex events with annotated caption sentences describing the events that occur. These events may occur over various periods of time and may co-occur. In contrast to existing methods \cite{HMN,MGRMP,SGN,SAAT}, our TextKG understands complex events more comprehensively with the help of exploiting information from transcripts and integrating the relevant knowledge in knowledge graph to enhance the common-sense information for video captioning.

\begin{table}[t]
\centering
\small{
\setlength{\tabcolsep}{11.0pt}
\begin{tabular}{c|cccc}
\toprule
Method &B &M &C & Rep \\
\midrule
HSE \cite{HSE} & 9.8 & 13.8 & 18.8 & 13.2 \\
GVD \cite{GVD} & 11.0 & 15.7 & 22.0 & 8.8 \\
GVDsup \cite{GVD} & \textbf{11.3} & 16.4 & 22.9 & 7.0 \\
Van-Trans \cite{MaskedTrans} & 9.8 & 15.6 & 22.2 & 7.8 \\
Trans-XL \cite{Transformer-XL} & 10.4 & 15.1 & 21.7 & 8.5 \\
Trans-XLRG \cite{Transformer-XLRG} & 10.2 & 14.8 & 20.4 & 8.9 \\
MART \cite{Mart} & 10.3 & 15.7 & 23.4 & \textbf{5.2} \\ \hline
TextKG & \textbf{11.3} & \textbf{16.5} & \textbf{26.6} & 6.3 \\
\bottomrule
\end{tabular}}
\caption{Evaluation results on the Activity-Net Captions \textit{ae-val} subset in the paragraph-level evaluation mode.}
\label{tab-activity-paragraph}
\vspace{-4mm}
\end{table}

{\noindent \textbf{MSR-VTT dataset}.} The evaluation results on the MSR-VTT dataset are reported in Table \ref{tab-comparison-msrvtt-micro}. For the fair comparison, we use the same visual features as the existing methods \cite{HMN, MGRMP,ORG-TRL,SAAT}. In contrast to existing methods focusing on network architecture design to exploit visual information, our TextKG aims to exploit mutli-modality information in external knowledge graph and original videos, achieving the state-of-the-art performance on $3$ out of $4$ metrics, \ie, BLEU, METEOR, and CIDEr.  

Meanwhile, we also compare the TextKG method to the SOTA methods focusing on multi-modality pretraining models in Table \ref{tab-comparison-mrsvtt-micro}. As shown in Table \ref{tab-comparison-mrsvtt-micro}, our TextKG method performs favorably against existing methods, such as DECEM \cite{DECEM}, UniVL \cite{univl} and MV-GPT \cite{MVC}, that are pretrained on the large-scale datasets (\eg, Howto100M \cite{howto100}), \ie, improve $0.7$ absolute CIDEr score compare to the SOTA method LAVENDER \cite{lavender}.

\begin{table}[t]
\centering
\setlength{\tabcolsep}{11.0pt}
\small{
\begin{tabular}{c|cccc}
\toprule
Method &B &M &R &C \\ \hline
OA-BTG \cite{OA-BTG} & 41.4 & 28.2 & - & 46.9 \\
POS-CG \cite{POS-CG} & 42 & 28.2 & 61.6 & 48.7 \\
MGSA \cite{MGSA} & 42.4 & 27.6 & - & 47.5 \\
STG-KD \cite{STG-KD} & 40.5 & 28.3 & 60.9 & 47.1 \\
ORG-TRL \cite{ORG-TRL} & 43.6 & 28.8 & 62.1 & 50.9 \\
SGN \cite{SGN} & 40.8 & 28.3 & 60.8 & 49.5 \\
MGRMP \cite{MGRMP} & 41.7 & 28.9 & 62.1 & 51.4 \\
HMN \cite{HMN} & 43.5 & 29 & \textbf{62.7} & 51.5 \\ \hline
TextKG & \textbf{43.7} & \textbf{29.6} & 62.4 & \textbf{52.4}  \\
\bottomrule
\end{tabular}}
\caption{Evaluation results on the MSR-VTT test subset in the micro-level evaluation mode.}
\label{tab-comparison-msrvtt-micro}
\vspace{-3mm}
\end{table}

\begin{table}[t]
\small
\centering
\setlength{\tabcolsep}{3.0pt}
\small{
\begin{tabular}{c|c|c|cccc}
\toprule
Method & Input & Features &B &M &R &C \\ \hline
SWINBERT \cite{Swinbert} & V & VidSwin & 45.4 & 30.6 & 64.1 & 55.9 \\
CLIP4C \cite{Clip4caption} & V & CLIP & 46.1 & 30.7 & 63.7 & 57.7 \\
CMVC \cite{cmvc} & V & CLIP & \textbf{48.2} & \textbf{31.3} & \textbf{64.8} & 58.7 \\ 
LAVENDER \cite{lavender} &V & VidSwin & - & - & - & 60.1 \\
DeCEM \cite{DECEM} & V+S & BERT & 45.2 & 29.7 & 64.7 & 52.3 \\
UniVL \cite{univl} & V+S & S3D & 41.8 & 28.9 & 60.8 & 50.0 \\
MV-GPT \cite{MVC} & V+S & ViViT & 48.9$^*$ & 38.7$^*$ & 64.0 & 60.0\\ \hline
TextKG (CLIP) & V+S & CLIP & 46.6 & 30.5 & \textbf{64.8} & \textbf{60.8} \\ 
\bottomrule
\end{tabular}}
\caption{Comparison to the methods focusing on model pretraining on the MSR-VTT dataset in the micro-level evaluation mode. Notably, following CLIP4C, we use the pre-trained CLIP model on LAION-400M \cite{DBLP:journals/corr/abs-2111-02114} to extract video appearance features. $\ast$ The authors use a different library to compute BLEU and METOR. Thus, the results on BLEU and METOR are not directly comparable to other methods.}
\label{tab-comparison-mrsvtt-micro}
\end{table}

{\noindent \textbf{MSVD dataset}.} We also evaluate the proposed TextKG method on the MSVD dataset \cite{DBLP:conf/acl/ChenD11} that doesn't have speech and repoprt the results in Table \ref{tab-comparison-msvd-micro}. As shown in Table \ref{tab-comparison-msvd-micro}, our TextKG achieves the best results without speech transcripts by improving $1.2\%$ and $3.8\%$ CIDEr scores compared to the SOTA method HMN \cite{HMN} and the JCRR method \cite{bad} exploiting knowledge graph to model the relations between objects, demonstrating the effectiveness of the knowledge graph usage in our method for video captioning. 

\begin{table}[t]
\centering
\setlength{\tabcolsep}{11.0pt}
\small{
\begin{tabular}{c|cccc}
\toprule
Method &B &M &R &C \\ \hline
OA-BTG \cite{OA-BTG} & 56.9 & 36.2 & - & 90.6 \\
POS-CG \cite{POS-CG} & 52.5 & 34.1 & 71.3 & 88.7 \\
MGSA \cite{MGSA} & 53.4 & 35 & - & 86.7 \\
STG-KD \cite{STG-KD} & 52.2 & 36.9 & 73.9 & 93.0 \\
ORG-TRL \cite{ORG-TRL} & 54.3 & 36.4 & 73.9 & 95.2 \\
SGN \cite{SGN} & 52.8 & 35.5 & 72.9 & 94.3 \\
MGRMP \cite{MGRMP} & 55.8 & 36.9 & 74.5 & 98.5 \\
JCRR \cite{bad} & 57.0 & 36.8 & - & 96.8 \\
HMN \cite{HMN} & 59.2 & 37.7 & \textbf{75.1} & 104.0 \\ \hline
TextKG & \textbf{60.8} & \textbf{38.5} & \textbf{75.1} & \textbf{105.2}  \\
\bottomrule
\end{tabular}}
\caption{Evaluation results on the MSVD test subset in the micro-level evaluation mode.}
\label{tab-comparison-msvd-micro}
\end{table}

\begin{table}[h]
\centering
\setlength{\tabcolsep}{3.5pt}
\small{
\begin{tabular}{ccc|ccc|cccc}
\hline
\begin{tabular}[c]{@{}c@{}}V-F\end{tabular} & \begin{tabular}[c]{@{}c@{}}R-F\end{tabular} & Text & 
\begin{tabular}[c]{@{}c@{}}G-KG\end{tabular} & 
\begin{tabular}[c]{@{}c@{}}S-KG\end{tabular} & \begin{tabular}[c]{@{}c@{}}K-S\end{tabular} &B &M &C &Rep \\ \hline
\checkmark & & & & & & 7.4 & 15.7 & 32.1 & 4.1 \\
\checkmark & \checkmark & & & & & 9.5 & 17.7 & 45.9 & 5.2 \\
\checkmark & \checkmark & & \checkmark & & \checkmark & 9.7 & 17.8 & 48.9 & 4.2 \\
\checkmark & \checkmark & & & \checkmark & \checkmark & 9.7 & 18.0 & 48.5 & 3.3 \\
\checkmark & \checkmark & & \checkmark & \checkmark & \checkmark & 9.6 & 17.7 & 49.8 & 5.5 \\\hline
\checkmark & & \checkmark & & & & 13.0 & 21.2 & 62.5 & 2.7 \\
\checkmark & \checkmark & \checkmark & & & & 13.9 & {\bf 22.1} & 71.3 & 2.9 \\
\checkmark & \checkmark & \checkmark & \checkmark & & \checkmark & 13.7 & 22.0 & 73.5 & {\bf 2.0} \\
\checkmark & \checkmark & \checkmark & & \checkmark & \checkmark & 13.5 & 21.9 & 74.8 & 2.1\\
\checkmark & \checkmark & \checkmark & \checkmark & \checkmark &  & 13.7 & 21.8 & 72.0 & 2.8\\
\checkmark & \checkmark & \checkmark & \checkmark & \checkmark & \checkmark & {\bf 14.0} & {\bf 22.1} & {\bf 75.9} & 2.8 \\
\hline
\end{tabular}}
\caption{Ablation study on the YouCookII val subset in the paragraph-level evaluation mode. ``V-F'' and ``R-F'' indicate the video and region features, ``Text'' indicates the speech transcripts features, ``G-KG'' and ``S-KG'' indicate general and specific knowledge graph, and ``K-S'' indicates the knowledge selection mechanism.}
\vspace{-5mm}
\label{tab-ablation-youcookii-paragraph}

\end{table}

\begin{figure*}[t]
 \centering
 \includegraphics[trim=42 210 90 40, clip, width=0.90\linewidth]{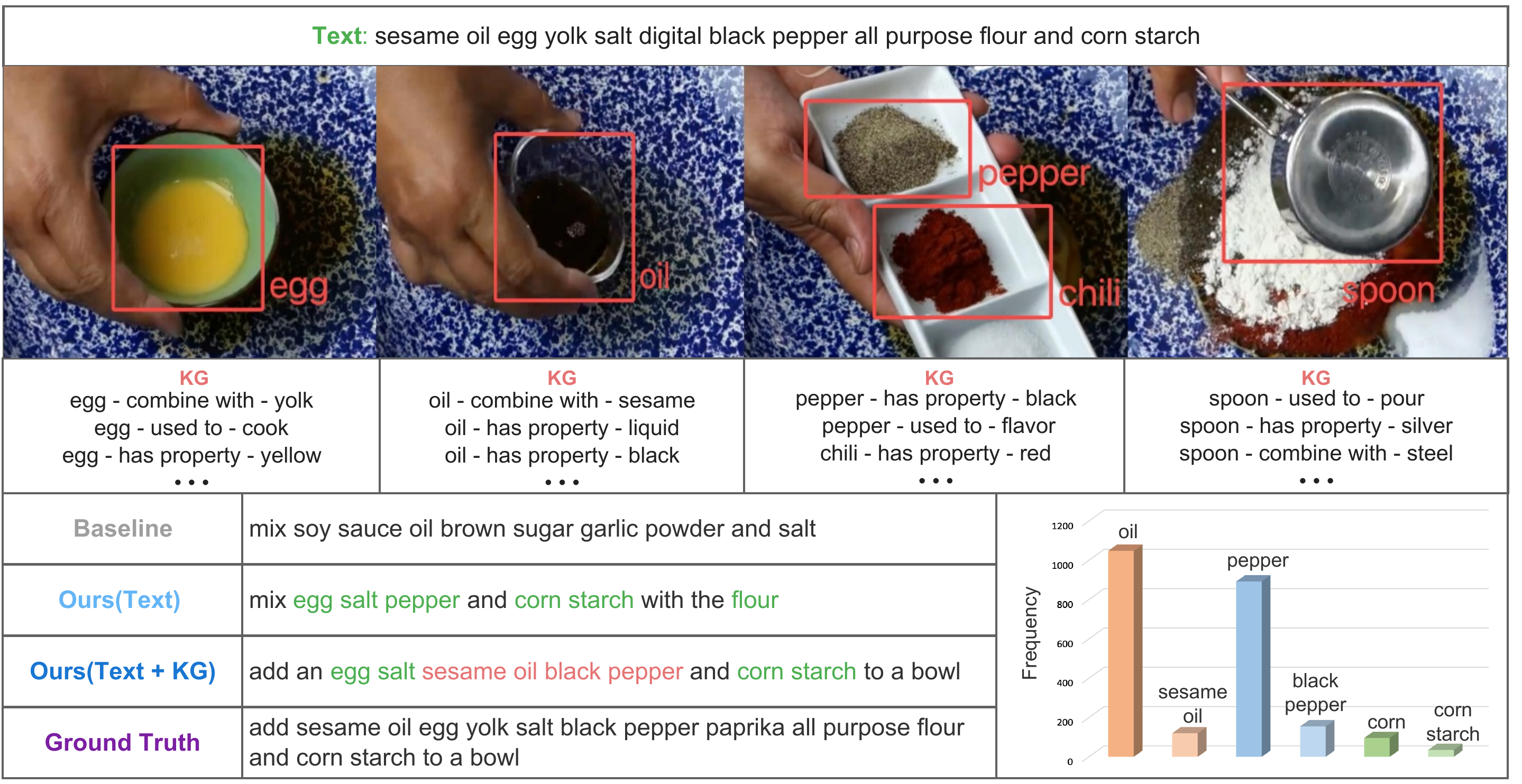}
 \caption{Qualitative results of the proposed method on the YouCookII dataset. The information in the speech transcripts and knowledge graph is not considered in the baseline.}
 \label{qualitative-results}
\end{figure*}

\subsection{Ablation study}
{\noindent {\bf Influence of different modules.}}
To validate the effectiveness of different modules in TextKG, we conduct several experiments on the YouCookII dataset in Table \ref{tab-ablation-youcookii-paragraph}. As shown in Table \ref{tab-ablation-youcookii-paragraph}, the baseline method considers the information of video features and predicted captions, achieving $32.1$ CIDEr score. After integrating the speech transcripts, the CIDEr score is improved to $62.5$, which ensures the key information of videos is considered in generating video captions. After integrating the region features, TextKG achieves $71.3$ CIDEr score. The CIDEr score can be further improved $4.6$ CIDEr score to $75.9$ by exploiting additional knowledge in knowledge graph. Notably, even without the information of speech transcripts, the knowledge graphs are still crucial for video captioning, \ie, improving $3.9$ CIDEr score ($49.8$ {\em vs.} $45.9$). Meanwhile, we also demonstrate the effectiveness of knowledge selection mechanism  in Table \ref{tab-ablation-youcookii-paragraph}. As shown in Table \ref{tab-ablation-youcookii-paragraph}, without the knowledge selection mechanism, the CIDEr score significantly drops $3.9$ to $72.0$. We believe that the noisy knowledge is harmful to the accuracy of video captions. 

\begin{figure}[t]
\centering
 \includegraphics[trim=20 0 40 30, clip, width=0.95\linewidth]{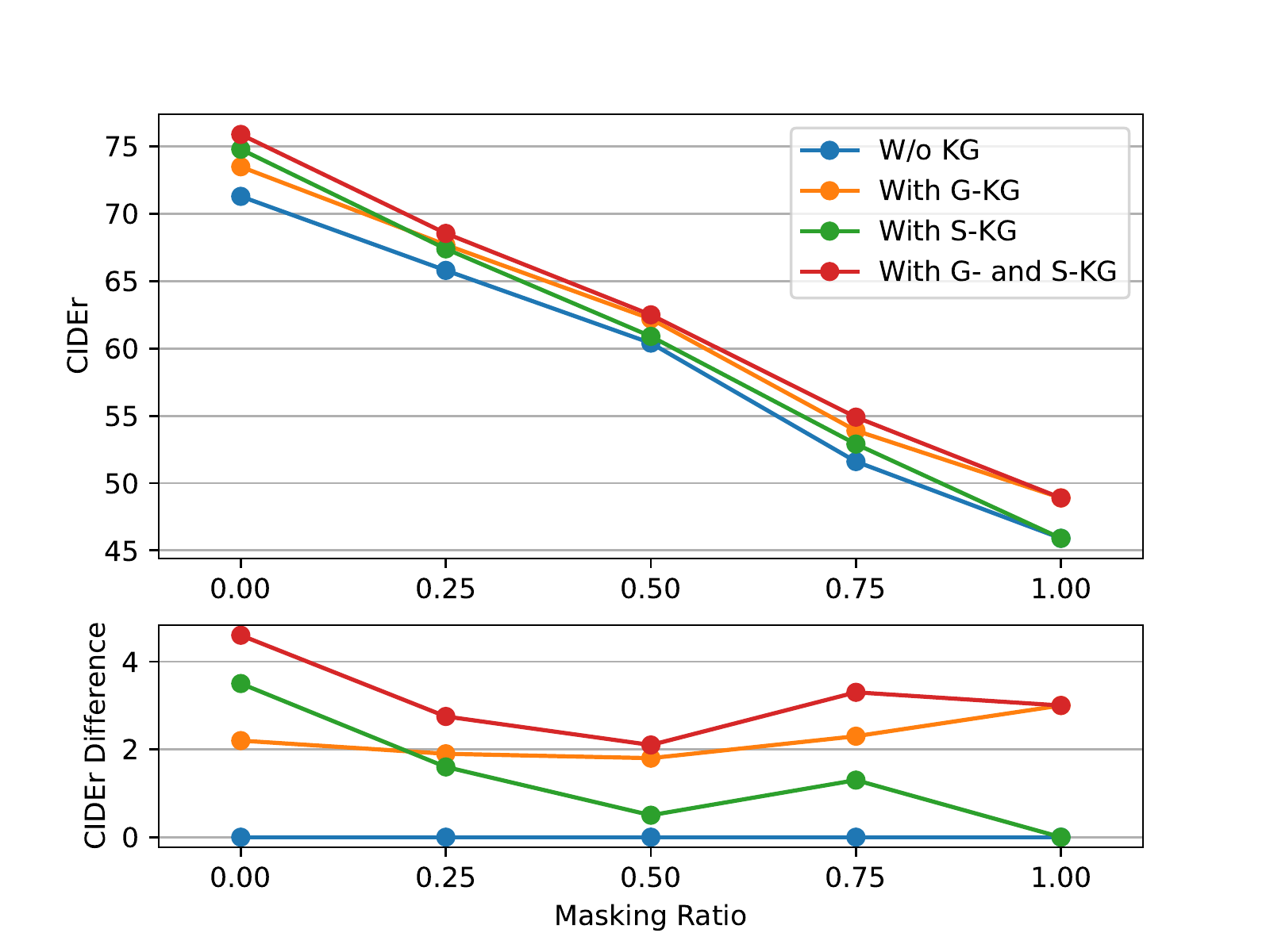}
 \caption{Influence of the transcripts qualities. ``w/o KG'' indicates that the method does not use the general and specific knowledge graphs. ``with G-KG'' and ``with S-KG'' indicate that only the general and specific knowledge graphs are used, respectively. ``with G- and S-KG'' indicates that both the general and specific knowledge graphs are used. The CIDEr differences between the ``with G-KG'', ``with S-KG'', ``with G- and S-KG'' methods and the ``w/o KG'' method are shown at the bottom corner.}
 \label{influence-transcripts}
 \vspace{-4mm}
\end{figure}

{\noindent {\bf Influence of transcript quality.}}
To explore the effect of transcript quality on the knowledge graph, we add some noises to the speech transcripts (\ie, mask some amount of speech transcripts) and evaluate the performance of our models on the YouCookII dataset in Figure \ref{influence-transcripts}. We find that the method using both the general and specific knowledge graphs performs favorable against other methods with different degrees of masked speech transcripts. Meanwhile, as the amount of masked speech transcripts increasing, the accuracy of the method using specific knowledge graph drops sharply, while the accuracy of the method using general knowledge graph drops at a much slower pace. The aforementioned results demonstrate the importance of general knowledge graph for the video captioning task.

\subsection{Qualitative Results}
We present the qualitative results of the proposed method on the YouCookII dataset in Figure \ref{qualitative-results}, including the key frames, detected salient objects, speech transcripts, retrieved knowledge items, caption results predicted by our model, ground-truth captions, and key words distributions in the training set. As shown in Figure \ref{qualitative-results}, we observe that the transcript contains key information in the video, \eg, egg, salt, pepper, com starch and flour, to ensure the quality of the generated video captions. However, the model with only the speech transcripts fails to predict the long-tail phrases, such as sesame oil and black pepper. With the help of knowledge graph providing the relations among these phrases, \eg, sesame oil, black pepper and corn starch, our TextKG model can predict the phrases more accurately. This indicates that the knowledge graph is crucial to mitigate the long-tail word challenges in video captioning.

\section{Conclusion}
In this paper, we present a text with knowledge graph augmented transformer for video captioning, which aims to integrate external knowledge in knowledge graph and exploit the multi-modality information in video to mitigate long-tail words challenge. Externsive experiments conducted on four challenging datasets demonstrate the effectiveness of the proposed method. 

In the future, we plan to improve the proposed method in two directions, \ie, (1) optimizing the knowledge retrieve strategy by considering semantic context information of detected objects and corresponding actions in videos; (2) constructing multi-modal knowledge graph (\eg, the nodes in knowledge graph formed by text, speech, and images or videos) to improve the accuracy of video captioning. 

{\small
\bibliographystyle{ieee_fullname}
\bibliography{refs}
}

\end{document}